\documentclass[10pt,twocolumn,letterpaper]{article}

\usepackage{btas}
\usepackage{times}
\usepackage{epsfig}
\usepackage{graphicx}
\usepackage{tabulary}
\usepackage{amsmath}
\usepackage{url}
\usepackage{multirow}
\usepackage{amssymb}
\graphicspath{{./graphics/}}

\renewcommand{\etal}{{\it et al.}~}

\btasfinalcopy 

\ifbtasfinal\fi
\begin{document}

\title{Thermal Features for Presentation Attack Detection in Hand Biometrics}

\author{Ewelina Bartuzi\\
Biometrics and Machine Intelligence Lab\\
Research and Academic Computer Network\\
Kolska 12, 01-145 Warsaw, Poland\\
{\tt\small ewelina.bartuzi@nask.pl}
\and
Mateusz Trokielewicz\\
Institute of Control and Computation Engineering\\
Warsaw University of Technology\\
Nowowiejska 15/19, 00-665 Warsaw, Poland\\ 
{\tt\small m.trokielewicz@elka.pw.edu.pl}
}

\maketitle
\thispagestyle{empty}

\begin{abstract}
This paper proposes a method for utilizing thermal features of the hand for the purpose of presentation attack detection (PAD) that can be employed in a hand biometrics system's pipeline. By envisaging two different operational modes of our system, and by employing a DCNN-based classifiers fine-tuned with a dataset of real and fake hand representations captured in both visible and thermal spectrum, we were able to bring two important deliverables. First, a PAD method operating in an open-set mode, capable of correctly discerning 100\% of fake thermal samples, achieving Attack Presentation Classification Error Rate (APCER) and Bona-Fide Presentation Classification Error Rate (BPCER) equal to 0\%, which can be easily implemented into any existing system as a separate component. Second, a hand biometrics system operating in a closed-set mode, that has PAD built right into the recognition pipeline, and operating simultaneously with the user-wise classification, achieving rank-1 recognition accuracy of up to 99.75\%. We also show that thermal images of the human hand, in addition to liveness features they carry, can also improve classification accuracy of a biometric system, when coupled with visible light images. To follow the reproducibility guidelines and to stimulate further research in this area, we share the trained model weights, source codes, and a newly created dataset of fake hand representations with interested researchers.  
\let\thefootnote\relax\footnote{Paper accepted for the BTAS 2018 Special Session on Image and Video Forensics In Biometrics, 22-25 Oct, 2018, Los Angeles, USA}
\end{abstract}

\section{Introduction}
\label{sec:intro}
Personal features of the hand have been employed for the authentication of humans since the early days of modern biometrics, in the form of fingerprint minutiae introduced as early as 1892 by Galton \cite{galton1892finger}, geometric features \cite{HandGeometryStasiak2006, HandGeometryYoruk2006}, palmprints \cite{PalmprintZhang2003, Pamlprint2017_Tiwari, PalmprintFei2018, PalmprintKumar2018}, and finger and hand vein patterns \cite{HandVein2006, HandVein2011, HandVein2017}. Recently, thermal features of the hand have gained some attention in the biometric community with the work of Bartuzi \etal \cite{EwelinkaHandIWBF2018}, showing that heat distributions carry discriminatory information and allow to build a biometric method based solely on the thermal hand representations. 

Imaging of the human hand for biometric purposes is rather straightforward, and in cases when only texture features, such as principal lines or minutiae of the palmprint are used, does not require a specialized equipment. However, what makes it easy to collect a biometric sample, usually also reveals a presentation attack (PA) vulnerability, which may involve presenting the system with a fake representation of the hand, such as a paper printout, prosthetics, displays, or using a genuine hand in a non-conformant scenario (\eg, use under coercion, or presentation improper enough to compromise the system).

Thus, an important piece in every well-designed biometric system's pipeline is a way to mitigate such attempts, \ie, presentation attack detection (PAD). Although thermal features of the hand have been used as cues for determining a person's identity, we are not aware of any PAD method that would employ such traits for counteracting fake representation attacks. This paper thus offers the following \textbf{contributions to the state-of-the-art:}
\begin{itemize}
	\item a presentation attack detection method employing thermal features, using a static image of the hand, based on a deep convolutional neural network (DCNN) model trained in both identity-driven (closed set) and authenticity-driven (open-set) approaches,
	\item a dataset of fake hand representations collected in the visible and thermal infrared wavelength rages, complementary to the existing dataset of genuine hand samples from the \emph{MobiBits} database,
	\item trained model weights and source codes for the offered solution.
\end{itemize}

Model weights, source codes and a complementary dataset of fake visible light and thermal infrared hand representations can be obtained at: \url{http://zbum.ia.pw.edu.pl/EN/node/46}.

This article is laid out as follows. In Sec. \ref{sec:related} we review the existing PAD methods for alleviating spoof attacks in palmprint and dorsal hand vein biometrics. Sec. \ref{sec:data} describes a subset of the \emph{MobiBits} database used for the purpose of this study and details the process of creating a complementary set of fake hand representations for training and evaluating our PAD method, which is introduced in Sec. \ref{sec:methods}. Finally, experimental results are reported and discussed in Sec. \ref{sec:results}, whereas Sec. \ref{sec:conclusions} provides relevant conclusions.

\section{Related work}
\label{sec:related}
Kanhangad and Kumar explored local binary pattern features for discerning genuine palmprint samples from paper printouts, reaching over 97\% accuracy \cite{KanhangadHandPAD2013}. Kanhangad \etal pointed to a high, almost 80\% risk of accepting fake palmprint samples as genuine ones using one of the existing academic methods, and proposed an anti-spoofing mechanism that analyzes the reflectance of the palmprint samples, and is said to be able to correctly classify 99\% of the genuine samples, paper printouts, and computer display photographs into either \emph{real} or \emph{fake} subsets \cite{KanhangadHandPAD2015}. A method for detecting presentation attacks in dorsal hand-vein biometrics is introduced by Bhilare \etal in \cite{BhilareHandPAD2017}, employing a histogram of oriented gradients performed on LoG-filtered images and an SVM with majority voting for image classification, reaching EER from 0.16\% to 0.8\%. A fusion of texture-based approaches and image quality assessment for face and palmprint PAD is introduced by Farmanbar and Toygar in \cite{FarmanbarHandPAD2017}, and tested on several publicly available datasets of face and palmprint samples. Bhilare \etal followed up on their earlier work in \cite{PADhand2018}, introduing a spoof sample database -- 'PALMspoof', and a PAD method that is said to outperform their earlier LBP-based approach by 12.73 percentage points in classification error rate. 

All of the methods reviewed above employ hand-crafted texture features or image statistics of visible light or near infrared images for determining PAD cues. However, to our knowledge, there are no prior papers or published research that would employ thermal imaging for the purpose of presentation attack detection of hand samples, despite this being perhaps the most natural choice for building a PAD method, as replicating the exact heat distribution of the human hand seems a dubious and difficult task for an attacker to perform. 

\section{Experimental data}
\label{sec:data}

\subsection{Dataset of visible light and thermal hand images}
For the purpose of this work we use a subset of the \emph{MobiBits}, a multimodal mobile biometric database including  images of palm side of the hand \cite{Mobibits}. The data were collected from 53 volunteers (20 female and 33 male, both hands, bringing a total of 106 classes). The age of subjects ranged from 14 to 71 years. The data were collected in three different acquisition sessions organized during three following months, twice in each session, approximately 15 minutes apart. All samples were acquired in typical office conditions with air conditioning on and ambient temperature set to $24^o$C. 

%

The discussed subset, containing {\bf hand images} can be categorized into two sample types:

\begin{itemize}
\item[1.] {\bf RGB} -- images taken with a rear camera of the CAT s60 mobile phone (480 $\times$ 640 pixels) in three sessions: with no temperature influence, after warming with a hot pillow, and after cooling with a cold compress. Additionally, measurement included unconstrained acquisition (raised hand) and acquisition stabilized by a glass stand.
\item[2.] {\bf TH} -- thermal images, taken simultaneously with the images in visible light, constituting the fourth layer of the image: RGB + TH, but with lower resolution of 240 $\times$ 320 pixels).
\end{itemize}

\subsection{Creating fake hand representations}
\begin{figure}[h!]
\centering
\includegraphics[width=1\linewidth]{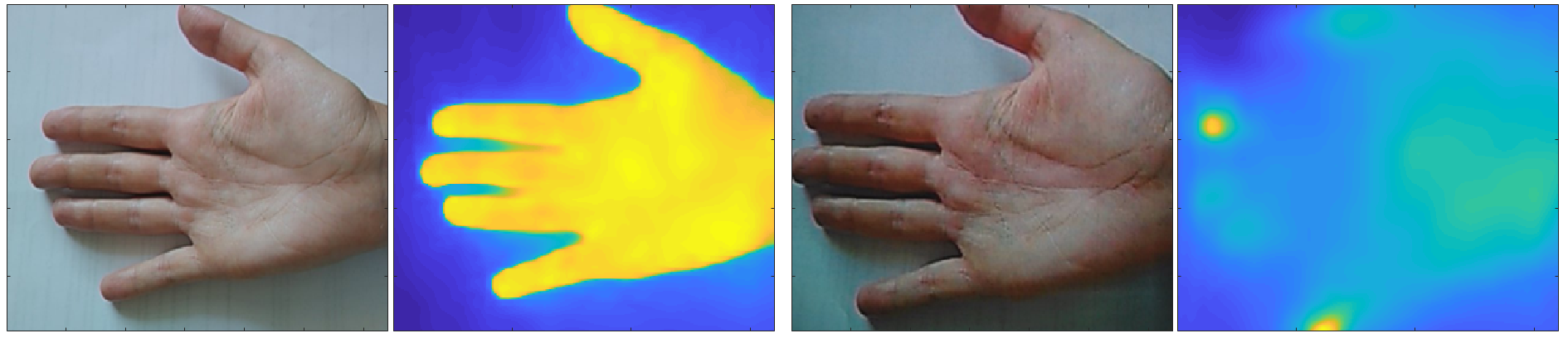}
\caption{{\bf From left to right:} an example real hand image in visible light, under thermal imaging, and respective fake representations.}
\label{creating_fakes}
\end{figure}

As a part of the experimentation performed in this work, we have extended the {\it MobiBits--Hand} subset with fake samples (two images per class). Hand images were printed and then photographed in a similar way as real hands were when creating the database, and with the use of a same CAT s60 device. Heat distribution of hand was imitated by the hand of a living human placed under the printout during data collection to make the presentation attack more plausible. Example hand images from the MobiBits-Hand subset together with their respective fake representations are shown in Fig. \ref{creating_fakes}. 

\section{Proposed methodology}
\label{sec:methods}

\subsection{Model architecture}
Two popular DCNN architectures were used in our experiments, namely the VGG-19 model introduced by Simonyan \cite{VGGSimonyanCNNsForRecognition2014}, and a shallower AlexNet model proposed by Krizhevsky \cite{AlexNet2014}. VGG-19, called 'very deep' at the time of publication, consists of 16 convolutional layers -- which serve as feature extractor, followed by three fully connected (FC) layers, constituting a classifier. AlexNet, on the other hand, comprises only 5 convolutional layers, followed by a similar, three-layer fully-connected classifier. Both networks participated in the ILSVRC competition \cite{ILSVRC} and are pre-trained on the ImageNet database \cite{imagenet} for the task of classifying natural images. 

\begin{table*}[!t]
\renewcommand{\arraystretch}{1.2}
\caption{Summary of the experimental protocol in each scenario (SPE - seconds per epoch).}
\label{table:trainingDetails}
\centering
\begin{tabular}{ |>{\centering\arraybackslash }m{0.2\linewidth}|>{\centering\arraybackslash }m{0.27\linewidth}|>{\centering\arraybackslash}m{0.27\linewidth}|} 
\hline
\multirow{2}{*}{{\bf Mode / DCNN model}} & \multirow{2}{*}{\textbf{AlexNet}} & \multirow{2}{*}{\textbf{VGG-19}}  \\
& & \\
\hline
\hline
\multirow{6}{*}{\textbf{Authenticity-driven}} & \multicolumn{2}{c} {\bf binary classification} \vline \\
& \multicolumn{2}{c}{{\bf open-set} (subject-disjoint)} \vline \\
& \multicolumn{2}{c}{{\bf 2 images per subject}} \vline \\
& \multicolumn{2}{c}{200 real  and 200 fake images} \vline \\

\cline{2-3} 
 	&  RGB: 12  epochs &   RGB: 10 epochs \\
	&  TH: 11  epochs  &   TH: 7 epochs \\
\hline
\multirow{6}{*}{\textbf{Identity-driven}} & \multicolumn{2}{c} {\bf class-wise prediction} \vline \\
& \multicolumn{2}{c}{106 identity classes $+$ 1 class of fake representations} \vline \\
& \multicolumn{2}{c}{{\bf closed-set} (sample-disjoint)} \vline \\
 & \multicolumn{2}{c}{{\bf 45 images per class} } \vline \\
\cline{2-3}
&  RGB:  45 epochs  & RGB: 25 epochs \\
&  TH:  75 epochs  & TH: 36 epochs \\
\hline
\end{tabular}
\end{table*}

These pre-trained models were then altered by modifying the bottleneck layers of the classifier stage to fit the task investigated in this paper, and fine-tuned with a database of genuine and fake, visible light and thermal hand images. These experiments were performed twofold, namely in an {\bf authenticity-driven open-set} scenario, and {\bf identity-driven closed-set} scenario. These are described in detail in the following sections.

%
 
\subsection{Authenticity-driven mode}
First, the DCNNs were used as binary classifiers, yielding a decision related to the authenticity of the sample being processed by the network, regardless of its claimed identity. The result here is assigning the sample either a \emph{real} or \emph{fake} label, together with a probability score obtained from one of the two output softmax neurons of the model. 

The DCNN operating in this mode can serve as an additional security element in any biometric system's pipeline, without introducing modifications to its architecture.  

\subsection{Identity-driven mode}
In the second part of the experiments reported in this paper, the DCNNs were fitted to operate in a closed-set scenario, where detection of fake hand representations is carried out simultaneously with providing a class-wise prediction of the probe sample. The last FC layers of each model were modified so that the number of output softmax neurons is $N+1$, where $N$ is the number of classes, and the additional neuron represents the \emph{fake} class. If a sample is classified as genuine, a typical prediction of the target class is given, together with a probability score obtained from the softmax layer. If, however, a fake is detected, then the sample is assigned a \emph{fake} label with a probability score.  

For a biometric system expected to operate in a closed-set mode only, this simplifies the integration of the PAD component into the recognition pipeline with little additional cost.   

\subsection{Training and evaluation}
For the network training and testing procedure, 10 subject-disjoint train/validation/test data splits were created. They were made with replacement, making them statistically independent and allowing us to assess the variance of the estimated error rates. The network was then trained with each train subset independently for each split, with the training being automatically stopped after achieving a non-increasing accuracy on the validation subset with patience of 10, and evaluated on the corresponding test subset. The training was performed with stochastic gradient descent as the minimization method with momentum $m=0.9$ and learning rate of 0.0001, with the data being passed through the network in mini batches of 16 images. Additionally, the training data were shuffled before each training epoch.

For the authenticity-driven, open-set mode, the data were divided in a subject-disjoint manner, so that classes chosen randomly for each of the train/validation/test splits do not overlap subject-wise. For the identity driven closed-set mode, subject-disjoint division is not possible, therefore for the second experiment the samples for each of the train/validation/test subsets were chosen randomly from each class to approximately meet the 60:20:20 criterion. Table \ref{table:trainingDetails} summarizes details of the experiments performed for each network and each scenario.  

In each mode, the networks were trained separately with RGB (visible light) and TH (thermal) images. Then, a score fusion at the softmax level was performed, so that in each mode three score distributions for each DCNN model were available: 

\begin{itemize}
	\item scores obtained from RGB images only,
	\vspace*{-2mm}
	\item scores obtained from TH images only,
	\vspace*{-2mm}
	\item scores obtained by averaging the above scores.
\end{itemize}

\section{Results and discussion}
\label{sec:results}

\subsection{Authenticity-driven mode}
\begin{figure}[h!]
	\centering
	\includegraphics[width=0.48\textwidth]{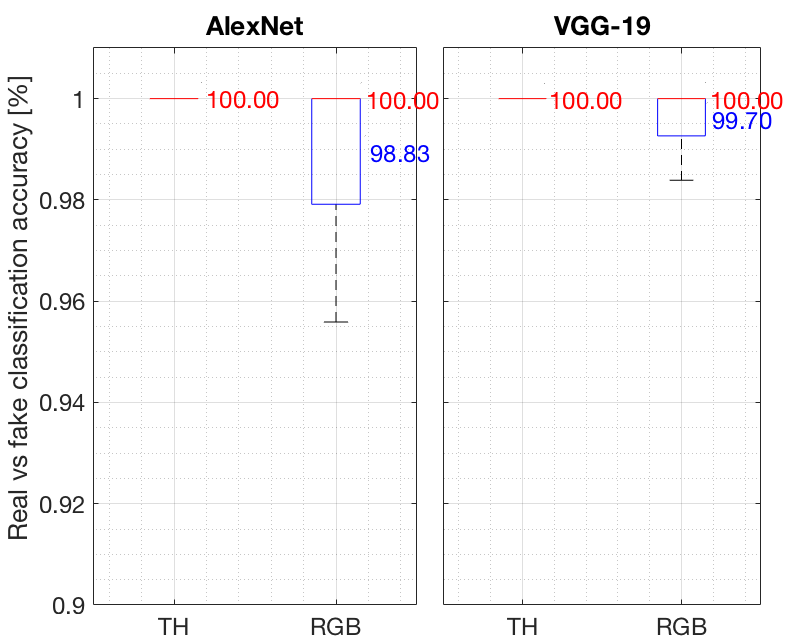}
	\caption{Boxplots representing differences in accuracy of classification into real and fake classes for thermal (TH) and visible light (RGB) hand representations for two DCNN models. Median values are shown in red, whereas mean values are shown in blue. Height of each box denotes an inter-quartile range (IQR), spanning from the first (Q1) to the third (Q3) quartile, whereas whiskers span from Q1-1.5*IQR to Q3+1.5*IQR.}
	\label{fig:authenticity-boxplots}
\end{figure}

Figure \ref{fig:authenticity-boxplots} presents the accuracy of classifying hand representations into either the \emph{real} or the \emph{fake} class, for both the AlexNet and VGG-19 models. Both classifiers achieve perfect accuracy when confronted with thermal samples, but even with only the visible light samples, we can still expect a very high, more than 98\% classification accuracy provided by the larger VGG-19, and slightly less than 96\% obtained from the much shallower AlexNet model. Since employing thermal features yield a perfect accuracy alone, we do not report on the score fusion results here. 

\subsection{Identity-driven mode}
\begin{figure}[h!]
	\centering
	\includegraphics[width=0.48\textwidth]{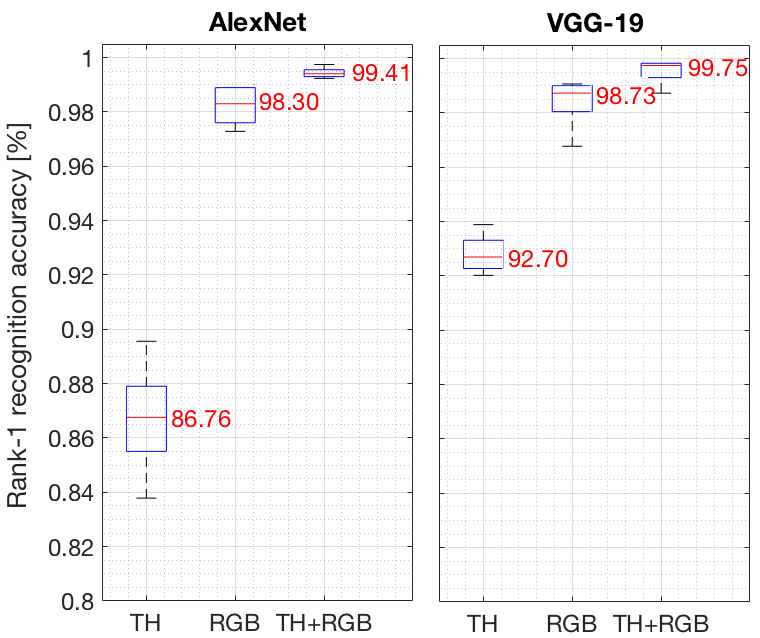}
	\caption{Same as in Fig. \ref{fig:authenticity-boxplots}, but showing accuracies for the identity-driven operational mode on a closed set of \emph{real} identities complemented by a \emph{fake} class. Scores obtained from thermal (TH) and visible light (RGB) hand representations, and scores obtained by averaging the TH and RGB scores (TH+RGB).}
	\label{fig:identity-boxplots}
\end{figure}

Classification accuracies obtained from the experiments performed using the identity-driven operational mode, in which the networks were given the task of not only detecting \emph{fake} representations, but also classifying the hand sample into one of the \emph{real} classes in a closed-set scenario, are shown in Fig. \ref{fig:identity-boxplots}.

\begin{figure}[t!]
	\centering
	\includegraphics[width=0.52\textwidth]{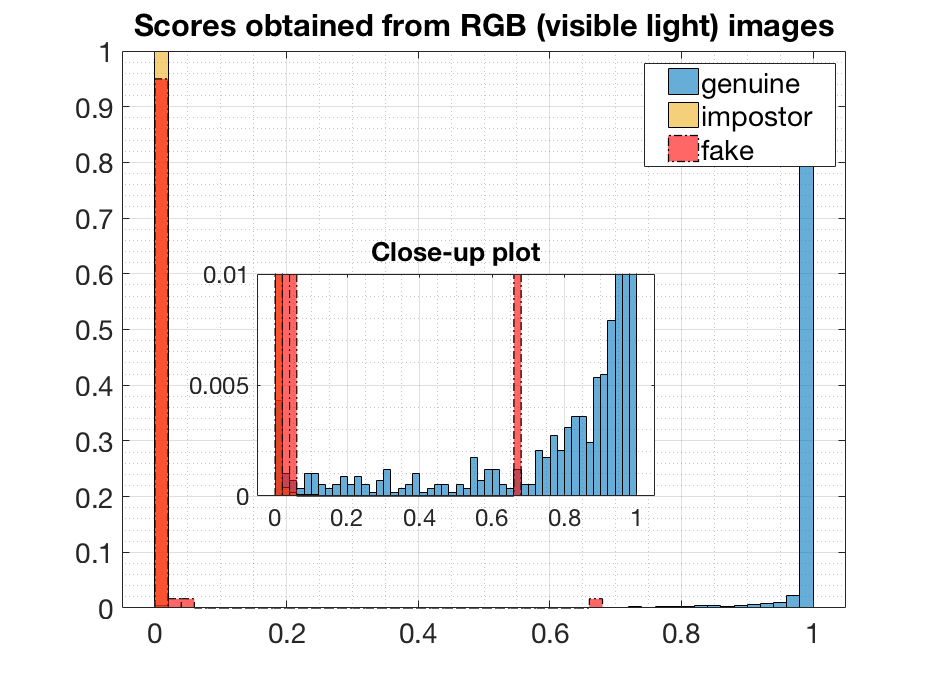}\\
	\includegraphics[width=0.52\textwidth]{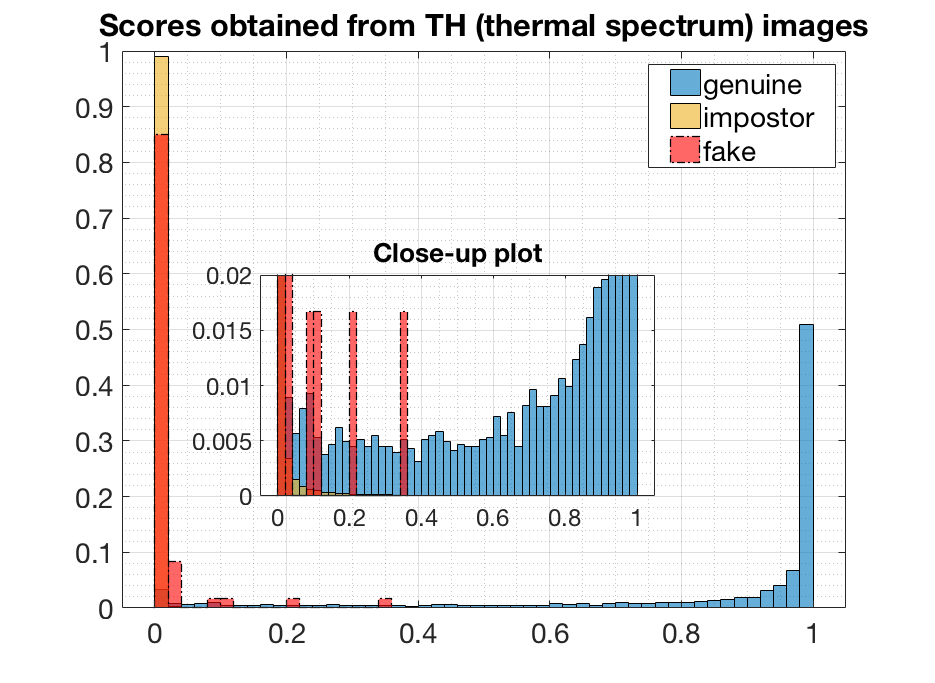}\\
	\includegraphics[width=0.52\textwidth]{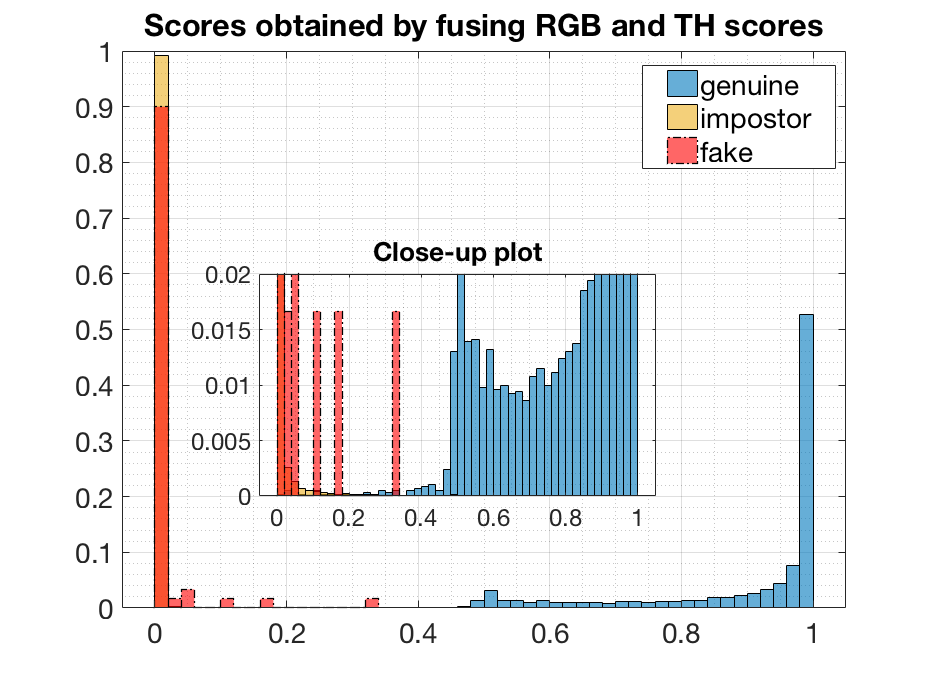}
	\caption{Score distributions generated with the AlexNet model for visible light images, thermal images, and a fusion of both. Smaller plots in the middle of each larger plot is a close-up of the lower registers of the ordinate axis.}
	\label{fig:identity-barplots-alexnet}
\end{figure}

\begin{figure}[t!]
	\centering
	\includegraphics[width=0.52\textwidth]{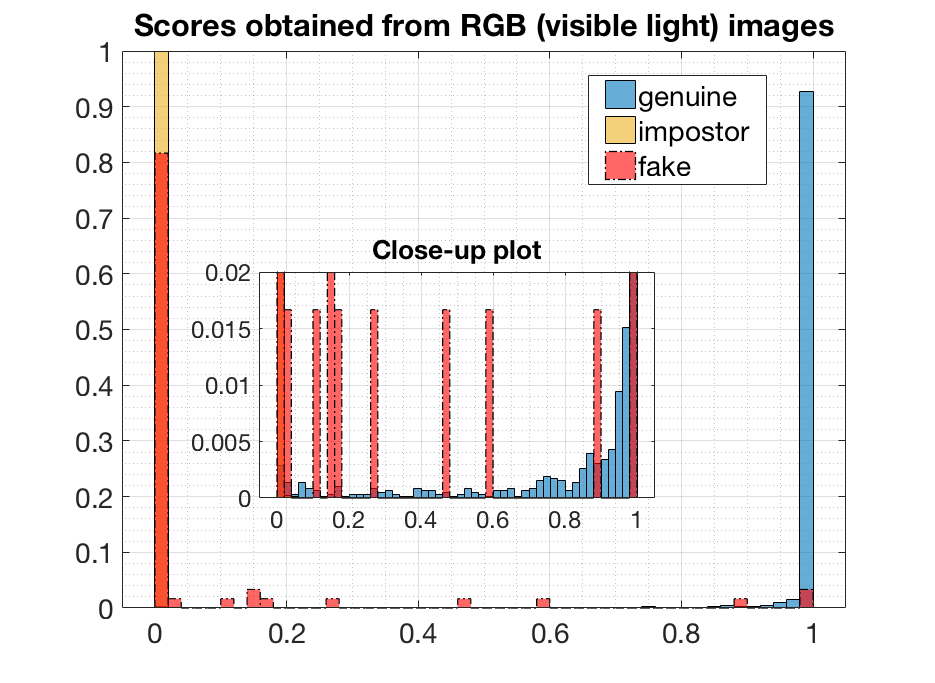}\\
	\includegraphics[width=0.52\textwidth]{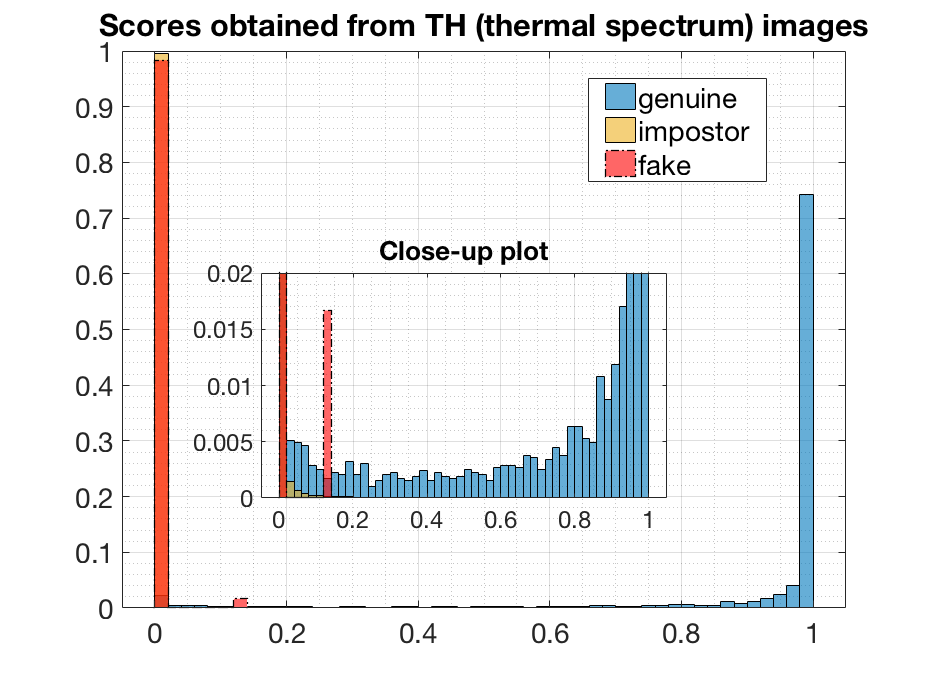}\\
	\includegraphics[width=0.52\textwidth]{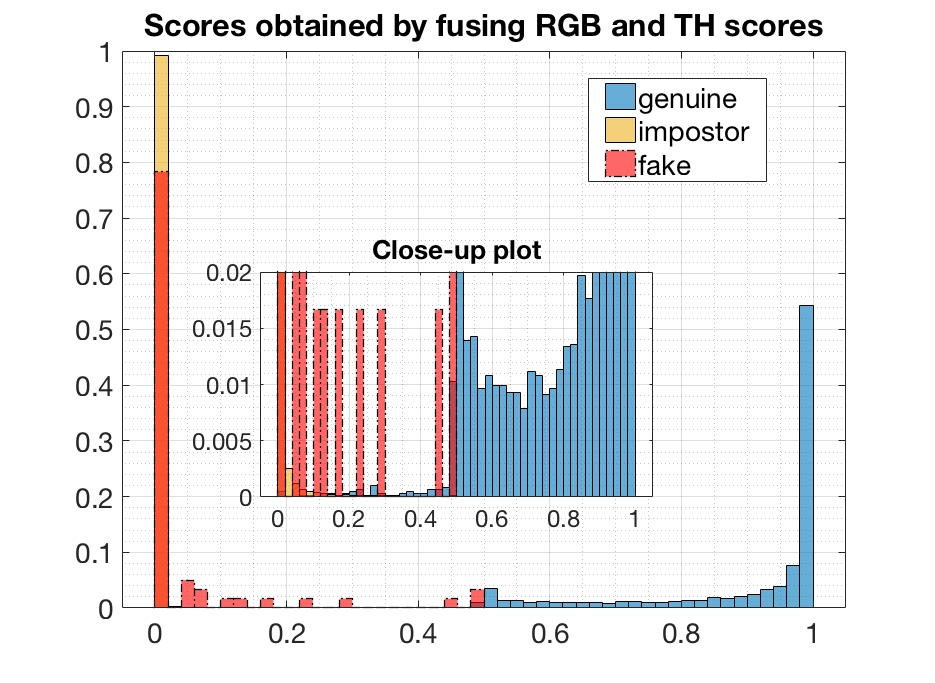}
	\caption{Same as in Fig. \ref{fig:identity-barplots-alexnet}, but for the VGG-19 model}
	\label{fig:identity-barplots-vgg19}
\end{figure}

In addition to classification accuracies, we also report the distributions of the scores used to generate boxplots shown in Fig. \ref{fig:identity-boxplots}. These are shown in Figures \ref{fig:identity-barplots-alexnet} and \ref{fig:identity-barplots-vgg19}, for scores obtained using the AlexNet and VGG-19 models, respectively. Notably, whereas some overlapping of the bins denoting scores corresponding to \emph{genuine} and \emph{fake} comparisons when RGB and TH images are used separately, this is almost completely rectified when a score fusion by arithmetical averaging is performed.

\section{Conclusions}
\label{sec:conclusions}
This paper is the first known to us work that explores thermal features of the hand for the purpose of assessing sample liveness for presentation attack detection, and employs deep convolutional neural networks adapted to two different operational scenarios and fine-tuned with a dataset of real and fake representations of hands imaged in both visible light and thermal spectrum. 

For the authenticity-driven mode, in which the algorithm's task is to discern \emph{fake} representations from \emph{real} ones, we were able to achieve perfect, 100\% accuracy averaged over 10 subject-disjoint, statistically independent train/validation/test data splits when thermal images are used with both the AlexNet and VGG-19 models. This translates to $APCER=0\%$ and $BPCER=0\%$. Also, surprisingly good performance can be expected for visible light images as well, with mean \emph{real} vs \emph{fake} classification accuracies equaling 98.83\% and 99.70\% for the AlexNet and VGG-19, respectively, averaged over the same training and testing procedure. These in turn allow to obtain low $APCER=0.87\%$ and $BPCER=0.55\%$ for the AlexNet, and $APCER=0.29\%$ and $BPCER=0.97\%$ for VGG-19. Since the much shallower AlexNet model reaches the same perfect performance with thermal images, we may hazard a guess that further optimization of the network architecture can bring down the computational cost, while maintaining the same 100\% accuracy. 

In the second operational scenario, namely the identity-driven closed-set mode, the solutions presented in this paper were able to offer 99.41\% and 99.75\% rank-1 recognition accuracy on the set of identities joined by a class of fake representations, for AlexNet and VGG-19, respectively. Although the VGG-19 model allows for a slightly better accuracy than AlexNet,  it is also much more computationally complex. Therefore, here as well we may argue that with further architecture optimizations, most of the recognition performance accuracy can be retained, while significantly reducing the cost of the solution. 

A second interesting observation that this experiment delivers, is that thermal features alone carry enough personal information to allow for over 92\% rank-1 recognition accuracy, and, when coupled with visible light features, are able to raise the overall performance of the system to an almost ideal accuracy of 99.75\%. Thus, \textbf{utilizing thermal features of the human hand appears not only to be a perfect method for a robust presentation attack detection method, but also a way to improve the overall performance of the biometrics system}, provided that both types of data are collected at both the enrollment, and the verification stages.

This work follows the IEEE guidelines for research reproducibility by offering the following contributions together with the paper: a) trained DCNN model weights and example source codes, and b) a dataset of fake hand representations that is complementary to the corresponding subset of the \emph{MobiBits} database \cite{Mobibits}.

{\small
\bibliographystyle{IEEEtran}
\bibliography{refs}
}

\end{document}